\def\year{2019}\relax
\documentclass[letterpaper]{article} 
\usepackage{aaai19}  
\usepackage{times}  
\usepackage{helvet}  
\usepackage{courier}  
\usepackage{url}            
\usepackage{graphicx}

\usepackage{amsfonts}       
\usepackage{amsmath}
\usepackage{multirow}
\usepackage{pifont}
\newcommand{\cmark}{\text{\ding{51}}}
\newcommand{\xmark}{\text{\ding{55}}}
\usepackage[noend]{algpseudocode}
\usepackage{algorithm}

\begin{document}
%
\title{A Layer Decomposition-Recomposition Framework for Neuron Pruning towards Accurate Lightweight Networks}
\author{Weijie Chen, Yuan Zhang, Di Xie, Shiliang Pu\\
Hikvision Research Institute\\
\{chenweijie5, zhangyuan, xiedi, pushiliang\}@hikvision.com}
\maketitle
\begin{abstract}
Neuron pruning is an efficient method to compress the network into a slimmer one for reducing the computational cost and storage overhead. Most of state-of-the-art results are obtained in a layer-by-layer optimization mode. It discards the unimportant input neurons and uses the survived ones to reconstruct the output neurons approaching to the original ones in a layer-by-layer manner. However, an unnoticed problem arises that the information loss is accumulated as layer increases since the survived neurons still do not encode the entire information as before. A better alternative is to propagate the entire useful information to reconstruct the pruned layer instead of directly discarding the less important neurons. To this end, we propose a novel \emph{Layer Decomposition-Recomposition Framework} (LDRF) for neuron pruning, by which each layer's output information is recovered in an embedding space and then propagated to reconstruct the following pruned layers with useful information preserved. We mainly conduct our experiments on ILSVRC-12 benchmark with VGG-16 and ResNet-50. What should be emphasized is that our results before end-to-end fine-tuning are significantly superior owing to the information-preserving property of our proposed framework. With end-to-end fine-tuning, we achieve state-of-the-art results of $5.13\times$ and $3\times$ speed-up with only 0.5\% and 0.65\% top-5 accuracy drop respectively, which outperform the existing neuron pruning methods.

\end{abstract}

\section{Introduction}
As the success of convolutional neural networks in computer vision, more and more researchers pay attention to their deployment on embedded sensors and mobile devices. Due to the enormous computational cost and storage overhead, it becomes an appealing subject on how to obtain a more efficient model without performance decline.

Recently, many compression and acceleration methods have been proposed to solve this problem. One category of the most popular methods is pruning, which can induce sparsity to weight matrices so that only the non-zero parts are involved in computation and storage. \cite{Guo2016Dynamic,Han2015Learning,Liu2015Sparse,Lebedev2016Fast,Wen2016Learning,Molchanov2016Pruning,Li2016Pruning,ye2018rethinking,yu2018nisp:}. 

\begin{figure}[ht]
\centering
\includegraphics[scale=0.39]{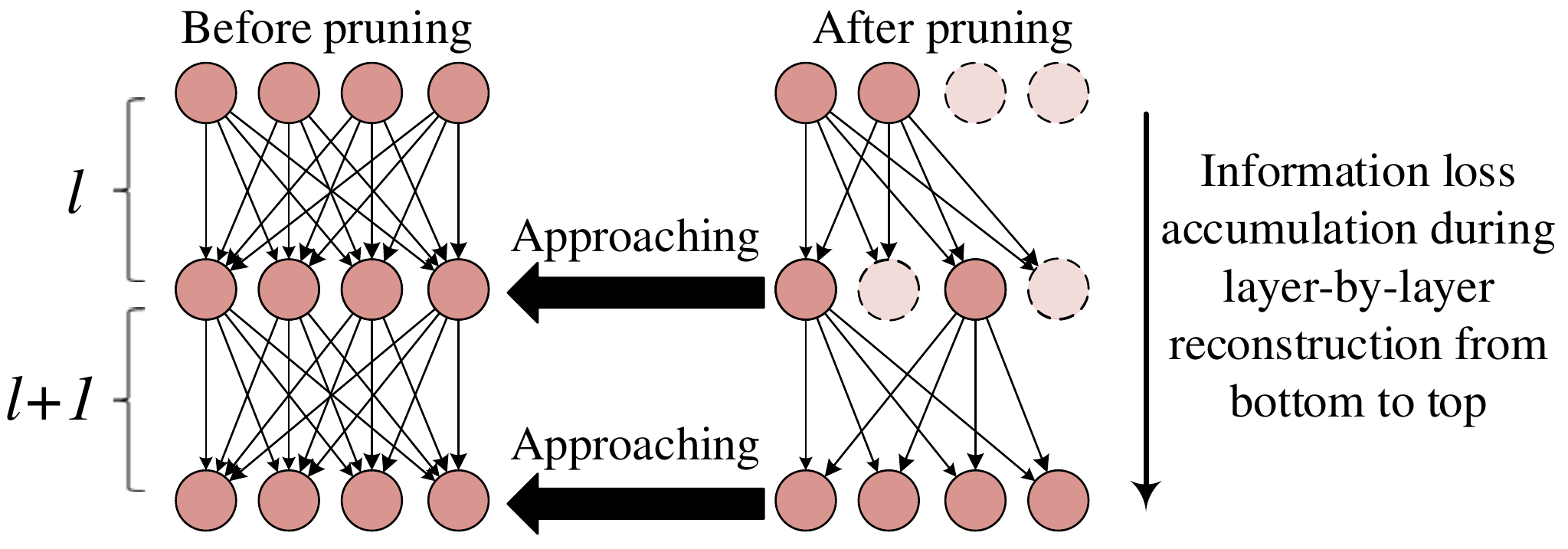}
\caption{The information loss of the generic layer-by-layer neuron pruning method. The less important input neurons in layer $l$ are pruned, and its output neurons are optimized to approximate the original ones. Repeatedly, only a part of important ones are propagated to reconstruct layer $l+1$, which induces information loss during the information propagation. Less importance does not mean absolute uselessness. A subset of neurons by selection is difficult to encode the entire label-related information especially when aggressive pruning. Such information loss is accumulated layer by layer, which deeply damages the network accuracy. For instance, \cite{He2017Channel} prune VGG-16 up to $5\times$ acceleration in such layer-by-layer pruning mode with {\bfseries 22\%} top-5 accuracy drop before end-to-end fine-tuning.}
\label{layerwise}
\end{figure}

Neuron pruning is one kind of structure pruning, which induces channel level and filter level sparsity of weight matrices so that it can directly compress the network into a slimmer one. To the best of our knowledge, most of state-of-the-art results are implemented in a layer-by-layer pruning and optimization manner \cite{Net-Trim,He2017Channel,Luo2017ThiNet}. However, as shown in Fig.\ref{layerwise}, the survived neurons are inadequate to recover the original output due to the information loss, which accumulates errors as layer increases. In this way, its accuracy is deeply damaged and is difficult to restore the original performance. \emph{In fact, this is an intrinsic drawback of the existing neuron pruning frameworks.} Their proposed greedy neuron selection methods actually attempt to ease this problem. However, without framework improved, this problem is unable to avoid, no matter how optimal the neuron selection is, especially when aggressive pruning.

Intuitively, a better solution is to squeeze the useful information of the discarded neurons into the survived ones and then propagate to reconstruct the following pruned layers. Inspired by this idea, we propose a new layer-by-layer optimization framework for neuron pruning named Layer Decomposition-Recomposition Framework (LDRF). In this framework, we elegantly generate the embedding space to each layer through cross-channel decomposition which serves as an information buffer. The information of the discarded neurons is squeezed into the survived ones in the embedding space before they are fed to reconstruct the following pruned layers. It efficiently avoids the emergence of information loss. Symmetrically, after pruning and reconstruction, each adjacent linear layers are recomposed to avoid network depth increase. Besides, via the analysis of the information propagation characteristic of Rectified Linear Units layer, i.e. ReLU \cite{dahl2013improving}, we propose a method to evaluate the pruning ratio range for each layer. It simplifies our optimization process and reduces our convergence time. We first conduct experiments on CIFAR-10 \cite{cifar} with VGG-9 \cite{VGG9} to compare our method with an existing layer-by-layer pruning method\cite{He2017Channel}. \emph{It is worth noting that the performance of pruned networks before end-to-end fine-tuning are the metrics to verify the effectiveness of such pruning methods.} Sadly, the majority of related works only compare the results after fine-tuning. In this experiment, the performance of our method significantly outperforms other methods before fine-tuning. Ultimately, we evaluate the generalization of our method on the large-scale ImageNet 2012 classification dataset \cite{Russakovsky2015ImageNet} with two most representative models, say VGG-16 \cite{Simonyan2014Very} and ResNet-50 \cite{He2016Deep}, and achieve state-of-the-art results by $5.13\times$ and $3\times$ speed-up with only {\bfseries 0.5\%} and {\bfseries 0.65\%} top-5 accuracy drop respectively. 

Our major contributions are summarized as follows:
\begin{itemize}
\item To the best of our knowledge, we are the first one to discover the hidden drawback of the existing layer-wise neuron pruning methods, i.e. the large information loss accumulation during the information propagation layer by layer, which leads to severe accuracy damage before end-to-end fine-tuning.
\item We propose a new and novel \emph{Layer Decomposition-Recomposition Framework} for neuron pruning to solve this problem effectively which preserves the useful information during the information propagation.
\item We achieve state-of-the-art results which outperform the existing methods. Additionally, before end-to-end fine-tuning, we have to emphasize that our results are significantly superior especially when aggressive pruning, which validates the impact of our proposed framework.
\end{itemize}

\section{Related Works}
{\bfseries Connection pruning} aims to prune unimportant connections and turns the dense weight matrices into sparse ones. As early as 1990s, Lecun \emph{et al.} \cite{lecun1990optimal} and Hassibi \emph{et al.} \cite{Hassibi1993Second} successively put forward optimal brain damage and optimal brain surgeon to remove useless parameters, so as to reduce model storage and computational overhead. Han \emph{et al.} \cite{Han2015Learning} propose a method to iteratively remove unimportant connections through $l_1$ or $l_2$ regularization. They claim that the connections with small weight below a given threshold are useless and it is harmless to the network accuracy if one removes them. However, it is difficult to accelerate in practice since the produced filter structures are irregular.

{\bfseries Structure pruning} is one kind of methods to prune filters into regular structure and brings convenience to practical deployment, especially on parallel computational platforms such as GPUs. Anwar \emph{et al.} \cite{anwar2017structured} adopt the particle filter to locate the unimportant region in channel level, kernel level and intra kernel dimensions. Li \emph{et al.} \cite{Li2016Pruning} simply use the evaluation criterion based on magnitude to prune unimportant filters. Molchanov \emph{et al.} \cite{Molchanov2016Pruning} compare different evaluation criteria for pruning and find that criterion based on Taylor expansion has superior performance among them. These works are tested in an end-to-end training manner. However, in a layer-by-layer pruning manner, a recent research \cite{randompruning} finds that the performances of pruned networks with different neuron importance evaluation criteria are comparable. They do several experiments and show that even the performance of random pruning is comparable with the one by other pruning criteria. This work inspires us that in a layer-by-layer pruning mode the core of pruning may lie in the capacity assignment to each layer (pruning ratio) which theoretically denotes the ceiling of the accuracy of the pruned networks instead of pruning criteria.

Group Lasso \cite{yuan2006model} is an efficient approach to learn the valid capacity of the network adaptively. Alvarez \emph{et al.} \cite{alvarez2016learning} introduce group lasso in the training process to learn the number of neurons in each layer. Wen \emph{et al.} \cite{Wen2016Learning} apply group lasso to exploit the redundancy in filters, channels, filter shapes, and network depth dimension. Besides, Liu \emph{et al.} \cite{slimming} take the scaling factor in BN layer to measure the importance of each channel and introduce a sparsity penalty on the scaling factors in the training process to learn the channel number. However, these methods are trained in an end-to-end manner and are difficult to recover the original accuracy. He \emph{et al.} \cite{He2017Channel}, Luo \emph{et al.} \cite{Luo2017ThiNet} and Aghasi \emph{et al.} \cite{Net-Trim} provide an analogous layer-wise optimization method to control the output consistency before and after pruning in each layer as shown in Fig. \ref{layerwise}, which can be formulated as:
\begin{equation}\label{0}
\min\limits_{W}\,\,\mathbb{g}{(W)}\quad\text{s.t.}\quad\parallel\textbf{y}-{\textbf{y}'}\parallel_{F}^{2}\leq\epsilon
\end{equation}
where $\textbf{y}$ and $\textbf{y}'$ correspond to the output in one layer before and after pruning respectively, $\epsilon$ is the affordable reconstruction error, and $\mathbb{g}(W)$ denotes group lasso or other sparsity-induced regularization of the layer's weight matrix $W$. 

Such layer-by-layer optimization achieves better performance compared with the end-to-end training manner, but it still has some problems on information propagation. As shown as Fig.\ref{layerwise}, there exists information loss during the reconstruction from shallow layers to high layers, since some less important but still informative neurons are discarded. This is the exact problem we want to solve in this paper.

\section{Our Approach}
\subsection{Pruning for Single Layer}\label{4.1}
Consider a single layer $\textbf{y}=\gamma(\textbf{W}^{T}\textbf{x})$, where $\textbf{x}\in\mathbb{R}^{m}$ is the layer input, $\textbf{W}\in\mathbb{R}^{m\times n}$ is a pretrained transformation matrix, $\textbf{y}\in\mathbb{R}^{n}$ is the output, and $\gamma(\cdot)$ is a popular non-linear activation function namely ReLU ($\max(\cdot  , 0)$) in this paper. Layer with this pattern is the basic unit of many popular networks, such as VGG-Net \cite{Simonyan2014Very}, GoogLeNet \cite{szegedy2015going} and ResNet \cite{He2016Deep}.

\begin{figure}[tp]
\centering
\includegraphics[scale=0.4]{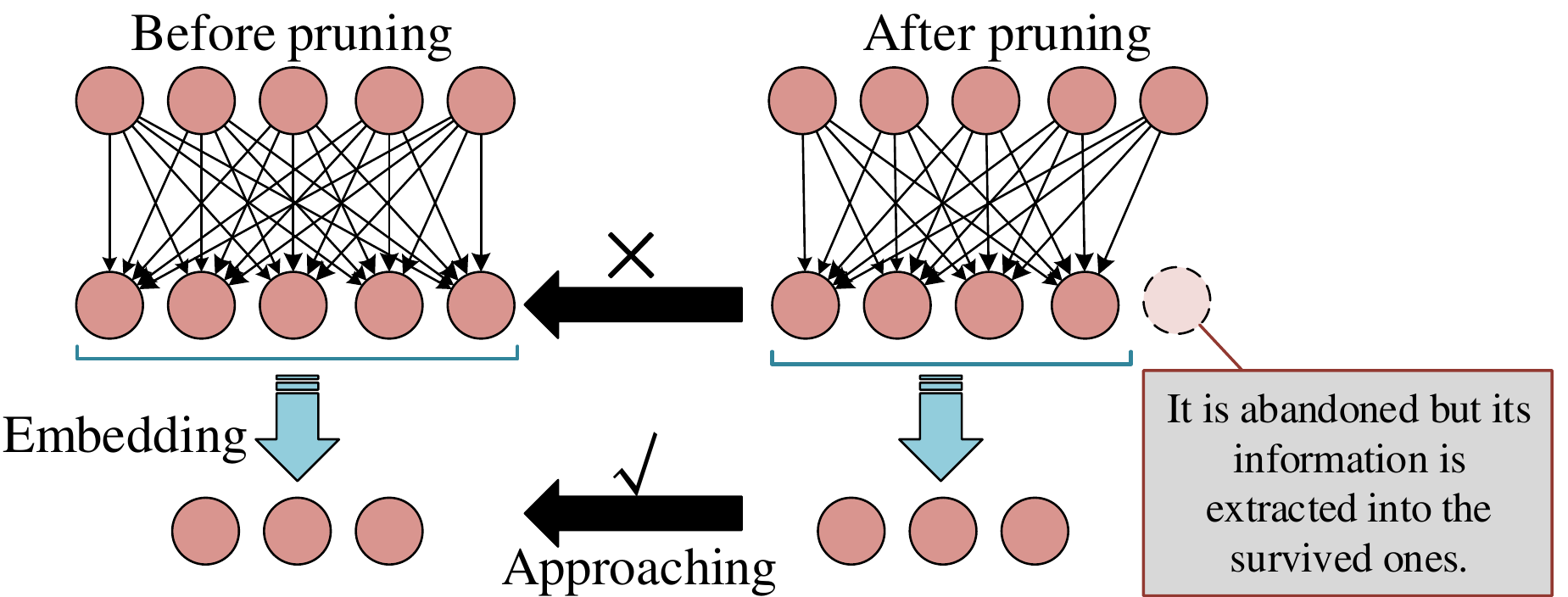}
\caption{Due to the output dimension inconsistency, the output neurons before and after pruning are projected into a shared embedding space for information compensation.}
\label{single_layer}
\end{figure}

In order to compress the layer, we aim to reduce output neurons while keeping the most information contained in the remaining neurons $\textbf{y}'\in\mathbb{R}^{n'}$. Here $\textbf{y}'=\gamma(\textbf{W}'^{T}\textbf{x})$ and $\textbf{W}'\in\mathbb{R}^{m\times n'}$ is a new transformation matrix with fewer columns ($n'\leq n$). As the dimension between $\textbf{y}$ and $\textbf{y}'$ is different, we suppose that the information of $\textbf{y}$ and $\textbf{y}'$ can be transformed into a shared embedding space, so that the difference can be measured which is helpful for compensating the information of discarded neurons, as shown in Fig. \ref{single_layer}. To this end, we formulate the neuron pruning problem as:
\begin{equation}\label{1}
\begin{aligned}
\min \limits_{0 < n' \leq n}n'\quad \text{s.t.}&\parallel\textbf{y}-(\textbf{Q}^{T})^{-1}\textbf{Q}^{T}\textbf{y}\parallel_{F}^{2}+ \\
&\lambda\parallel\textbf{Q}^{T}\textbf{y}-\textbf{Q}'^{T}\textbf{y}'\parallel_{F}^{2}\leq\epsilon
\end{aligned}
\end{equation}
where $\parallel\cdot\parallel_{F}$ is Frobenius norm, $\lambda$ is a weighted coefficient, and $\epsilon$ is our affordable information loss which is a sufficiently small scalar value. $\textbf{Q}\in\mathbb{R}^{n\times z}$ and $\textbf{Q}'\in\mathbb{R}^{n'\times z}$ in Eqn.\ref{1} are the projection matrices to embed $\textbf{y}$ and $\textbf{y}'$ into a comparable space while not losing useful information. The first term of Eqn.\ref{1} is to project the embedded representations back to original space with minimal error which makes sure that the embedding space is adequate to accommodate the entire information. However, Eqn.\ref{1} is difficult to solve because the shape of $\textbf{W}'$ and $\textbf{Q}'$ is changeable during optimization. To solve this problem, we fix the shape of $\textbf{W}'$ and $\textbf{Q}'$ as ones of $\textbf{W}$ and $\textbf{Q}$, and introduce a vector $\textbf{m}$ of size $n$ for neuron selection. Hence, $\textbf{y}'=\textbf{m}\odot\gamma(\textbf{W}'\textbf{x})$ where $\odot$ denotes point-wise multiplication. Note that the entries of $\textbf{m}$ are binary scalars. A $0$ means that the corresponding neuron is useless which can be pruned, while a $1$ indicates the expressive neuron. In this manner, the optimization problem to find the minimum of $n'$ is reduced to the problem to find the most sparse solution of $\textbf{m}$ ($l_0$ minimization problem). However, $l_0$ minimization problem is a NP-hard problem, so we relax $l_0$ to $l_1$ and convert Eqn.\ref{1} into the following optimization problem:
\begin{equation}\label{2}
\begin{aligned}
\mathop{\arg\min}_{\textbf{Q}, \textbf{Q}', \textbf{W}', \textbf{m}}&\parallel\textbf{y}-(\textbf{Q}^{T})^{-1}\textbf{Q}^{T}\textbf{y}\parallel_{F}^{2}+\lambda_{1}\parallel\textbf{Q}^{T}\textbf{y}-\\
&\textbf{Q}'^{T}(\textbf{m}\odot\gamma(\textbf{W}'^{T}\textbf{x}))\parallel_{F}^{2}+\lambda_{2}\parallel\textbf{m}\parallel_{1}
\end{aligned}
\end{equation}{}
where $\lambda_{1}$ and $\lambda_{2}$ are the weighted coefficients. Naturally, the projection matrix $\textbf{Q}$ can be obtained by Principal Component Analysis to a batch of $\textbf{y}$. Then, fixing $\textbf{Q}$ and initializing $\textbf{Q}'$ and $\textbf{W}'$ with $\textbf{Q}$ and $\textbf{W}$ to solve the remaining terms.

\subsection{Pruning for Entire Network}\label{4.2}
\begin{figure*}[!htbp]
\centering
\includegraphics[scale=0.5]{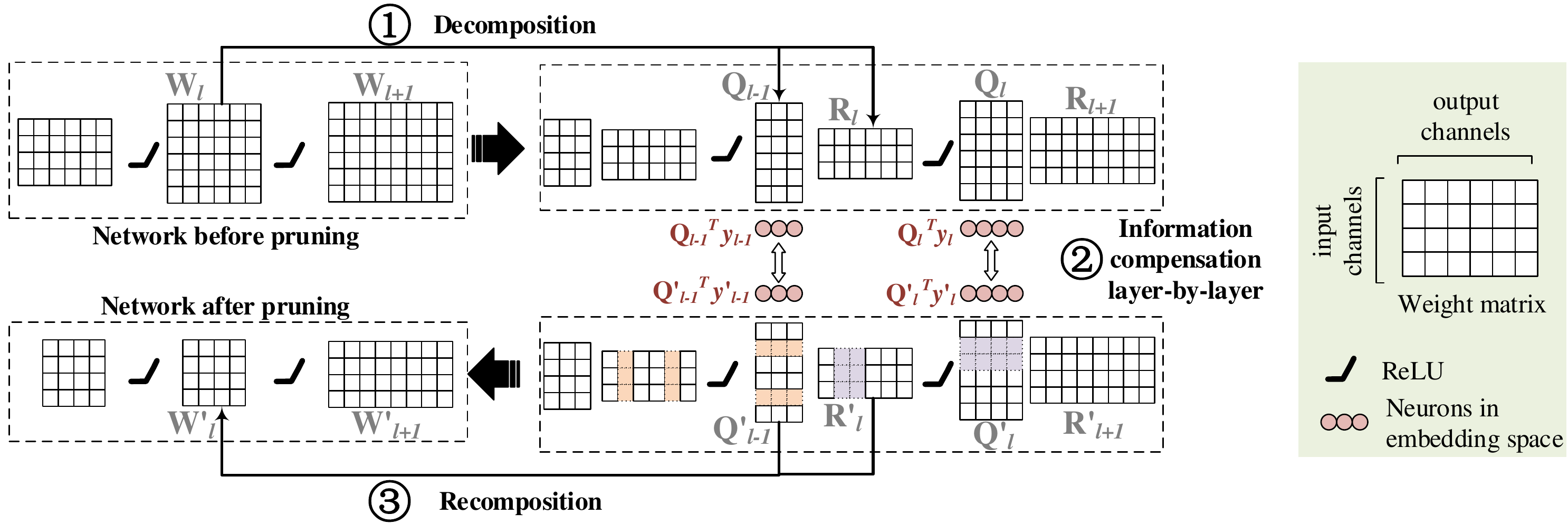}
\caption{A Layer Decomposition-Recomposition Framework for neuron pruning. Layer decomposition is first applied to each layer to generate its embedding space. Then the redundant neurons are pruned, which is equivalent to prune the output channels of $\textbf{R}'_{l}$ and the corresponding input channels of $\textbf{Q}'_{l}$, while compensating the information loss of discarded neurons in each embedding space. It means one should optimize the pruned $\textbf{R}'_{l}$ and $\textbf{Q}'_{l}$ to force ${\textbf{Q}'}_{l}^{T}\textbf{y}'_{l}$ to approximate $\textbf{Q}_{l}^{T}\textbf{y}_{l}$. Ultimately, a slimmer network without depth increasing is returned through layer recomposition. (Best viewed in color)}
\label{multi_layers}
\end{figure*}
In this section, the pruning method for one layer is generalized to the entire network in a layer-by-layer manner. $\textbf{y}_{l}=\gamma({\textbf{W}_{l}}^{T}\textbf{y}_{l-1})$ is the output of $l$th layer in a given pretrained network. ${\textbf{y}'}_{l}=\textbf{m}_{l}\odot\gamma({\textbf{W}'}_{l}^{T}{\textbf{y}'}_{l-1})$ is the compact output of $l$th pruned layer that attempts to produce similar information as $\textbf{y}_{l}$, where $\textbf{y}'_{l-1}$ is the generated output neurons in the $l$-$1$th layer which has been pruned. In line with pruning one layer, we aim to reduce the neurons of ${\textbf{y}'}_{l}$ while keeping its information comparable with $\textbf{y}_{l}$ in a shared embedding space. Thus our objective function is
\begin{equation}\label{3}
\begin{aligned}
\mathop{\arg\min}_{\textbf{Q}_{l}, \textbf{Q}'_{l}, \textbf{W}'_{l}, \textbf{m}_{l}}&\parallel\textbf{y}_{l}-(\textbf{Q}^{T}_{l})^{-1}\textbf{Q}^{T}_{l}\textbf{y}_{l}\parallel_{F}^{2}+\lambda_{1}\parallel\textbf{Q}^{T}_{l}\textbf{y}_{l}-\\
&{\textbf{Q}'}^{T}_{l}(\textbf{m}_{l}\odot\gamma({\textbf{W}'}_{l}^{T}\textbf{y}'_{l-1}))\parallel_{F}^{2}+\lambda_{2}\parallel\textbf{m}_{l}\parallel_{1}
\end{aligned}
\end{equation}
Different from pruning single layer, $\textbf{Q}'_{l}$ and $\textbf{W}'_{l}$ cannot be initialized with $\textbf{Q}_{l}$ and $\textbf{W}_{l}$ because the numerical representations of $\textbf{y}_{l-1}$ and $\textbf{y}'_{l-1}$ are not identical although they contain similar information\footnote{In single layer case, both input information is $\textbf{x}$, so they can inherit from $\textbf{Q}$ and $\textbf{W}$ safely.}. A better solution is to decompose $\textbf{Q}_{l-1}$ from $\textbf{W}_{l}$ ($\textbf{W}_{l}=\textbf{Q}_{l-1}\textbf{R}_{l}$). In this way, we can directly feed the embedding representation ${\textbf{Q}'}_{l-1}^{T}\textbf{y}'_{l-1}$ to the next layer with transformation matrix ${\textbf{R}'}_{l}$, where ${\textbf{R}'}_{l}$ and the following embedding matrix $\textbf{Q}'_{l}$ can be initialized with $\textbf{R}_{l}$ and $\textbf{Q}_{l}$. Moreover, in the situation of multi-layers, evaluating the expressiveness of $\textbf{Q}^{T}\textbf{y}$ is equivalent to evaluate whether it can flow to the next layer without information loss. So the first term of Eqn.\ref{3} is better to replace by $\parallel\textbf{W}_{l+1}^{T}\textbf{y}_{l}-\textbf{R}_{l+1}^{T}\textbf{Q}_{l}^T\textbf{y}_{l}\parallel_{F}^{2}$, and we can reformulate the objective function as follows:
\begin{equation}\label{4}
\begin{aligned}
\mathop{\arg\min}_{\textbf{Q}_{l}, \textbf{R}_{l+1}, {\textbf{Q}'}_{l}, {\textbf{R}'}_{l}, \textbf{m}_{l}}\parallel\textbf{W}_{l+1}^{T}\textbf{y}_{l}-\textbf{R}_{l+1}^{T}\textbf{Q}_{l}^T\textbf{y}_{l}\parallel_{F}^{2}+\lambda_{1}\parallel\textbf{Q}^{T}_{l}\textbf{y}_{l}-\\
{\textbf{Q}'}^{T}_{l}(\textbf{m}_{l}\odot\gamma({\textbf{R}'}_{l}^{T}({\textbf{Q}'}_{l-1}^{T}{\textbf{y}'}_{l-1})))\parallel_{F}^{2}+\lambda_{2}\parallel\textbf{m}_{l}\parallel_{1}
\end{aligned}
\end{equation}
$\textbf{R}_{l+1}$ and $\textbf{Q}_{l}$ in the first term can be determined by applying Singular Value Decomposition (SVD) to $\textbf{W}_{l+1}$. After that, given that $\textbf{R}_{l}$ and ${\textbf{Q}'}_{l-1}$ have been determined in the $l$-$1$th layer,  we initialize ${\textbf{Q}'}_{l}$ and ${\textbf{R}'}_{l}$ with $\textbf{Q}_{l}$ and $\textbf{R}_{l}$, and then solve ${\textbf{Q}'}_{l}$, ${\textbf{R}'}_{l}$ and $\textbf{m}_{l}$. Repeating this process, the entire network can be pruned in a layer-by-layer manner. According to the mask vector $\textbf{m}_{l}$ in each layer, we remove useless neurons and finally get a slimmer network. In order to avoid network depth increasing, we can get the new transformation matrix $\textbf{W}'_{l}$ for each layer by recomposing $\textbf{Q}'_{l-1}$ and $\textbf{R}'_{l}$ ($\textbf{W}'=\textbf{Q}'_{l-1}\textbf{R}'_{l}$). We summarize the entire procedures as a Layer Decomposition-Recomposition Framework for neuron pruning as shown in Fig. \ref{multi_layers}. From this figure, it is easy to see that the information of discarded neurons is recovered in the embedding space and is propagated to reconstruct the following pruned layers with information preserved.

\subsection{A More Efficient Solution}\label{4.3}
The optimization of Eqn.\ref{4} is divided into layer decomposition and information compensation for neuron pruning. However, in practice, solving the latter one is extremely time-consuming. Additionally, $\lambda_{2}$ in the third term of Eqn.\ref{4} is not only difficult to determine, but also empirical to some extent\footnote{For instance, if we want to speed the network up to $2\times$, we have no idea how to set an exact scalar value to $\lambda_{2}$ unless we run many trial-and-error experiments.}.
To deal with both problems, we propose a method to evaluate each layer's pruning ratio range under the guarantee of information preserved propagation.

The information propagation characteristic of ReLU is the prerequisite to inspire us to propose our method. It is a non-linear activation layer which collapses negative signals to zero and only allows positive signals through the layer. It may bring about information loss and hurt the expression capability of current layer. For irredundant signals through ReLU layer, it is inevitable to lose the flow-in information unless all input signals are positive. In this way ReLU degenerates to an identity transformation and loses its nonlinearity. To preserve the flow-in information and maintain the nonlinearity of ReLU, the Lemma 2 of the appendix in MobileNetV2 \cite{mobilenetv2} demonstrates that the irredundant input signals should be expanded to a higher dimensional representation for introducing artificial redundancies before they are fed into the ReLU layer so as to maintain the invertibility of ReLU. This analysis directly provides us an idea to estimate the range of valid neurons for each layer, and guides us to assign pruning ratio to each layer.

To achieve this, we pick out the first term of Eqn.\ref{4} and solve it for each layer at first before solving the remaining terms. In addition, we add a constrain to the first term to calculate the lowest dimensional space to accommodate flow-in information in each layer. So the first term becomes:
\begin{equation}\label{5}
\min \,\, z_{l} \quad \text{s.t.}\parallel\textbf{W}_{l}^{T}\textbf{y}_{l-1}-\textbf{R}_{l}^{T}(\textbf{Q}_{l-1}^T\textbf{y}_{l-1})\parallel_{F}^{2} \leq\epsilon
\end{equation}
where $\epsilon$ is our affordable information loss, and $\textbf{Q}_{l-1}^T\textbf{y}_{l-1}$ is an irredundant representation before it is fed into ReLU layer and its dimension is $z_{l}$. We can get singular values of $\textbf{W}_{l}$ by SVD decomposition and keep the cumulative sum of the largest ones as small as possible for rank reduction. Once the minimum of $z_{l}$ is determined, we can learn that $\parallel \textbf{m}_{l} \parallel_{0}>z_{l}$ is a necessary condition for the guarantee of lossless information propagation through the following ReLU layer, because only if $\textbf{R}_{l}$ acts as an expansion function can $\textbf{Q}_{l-1}^T\textbf{y}_{l-1}$ possibly passes through ReLU layer without information loss. This is the same motivation from which Sandler \emph{et al.} \cite{mobilenetv2} design inverted residual and linear bottlenecks for MobileNetV2, but from different views. Therefore, the number of preserving output neurons should at least range in $(z_{l}, n_{l}]$.

\begin{figure}[tp]
\centering
\includegraphics[scale=0.35]{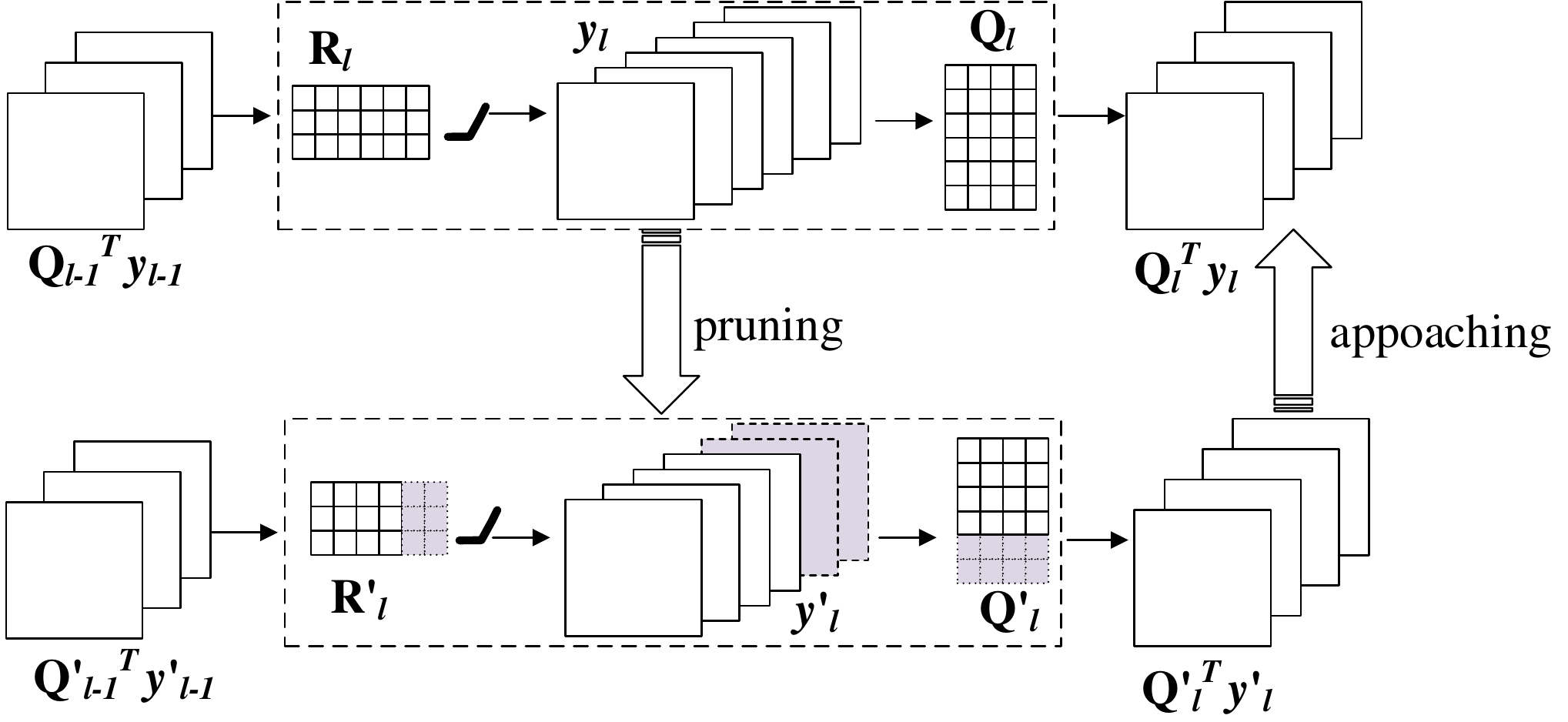}
\caption{Our framework propagates the entire information generated from the preceding layers to reconstruct current pruned layer WITHOUT input neuron (${\textbf{Q}'}_{l-1}^{T}{\textbf{y}'}_{l-1}$) selection. During the reconstruction, the information in the output embedding space (${\textbf{Q}'}_{l}^{T}{\textbf{y}'}_{l}$) is approaching to the original one, which indirectly squeeze the useful information of the intermediate pruned neurons into the survived ones (${\textbf{y}'}_{l}$).}
\label{basic_unit_analysis}
\end{figure}

Moreover, inspired by the experiments in \cite{randompruning} that even random pruning can provide a comparable result, it is the remaining capacity of each layer instead of a specific neuron selection criterion that matters to the final recovered network performance, on the premise of layer-by-layer training mode. In other words, we just concern about the amount of loss rather than which part of the information is selected to lose. Specific to our framework, as shown in Fig.\ref{basic_unit_analysis}, our framework is insensitive to neuron selection method since the intermediate pruned information is easy to restore in the output embedding space owing to the complete input information. So we simply use Top-k to set $\textbf{m}_{l}=[1_{1}, 1_{2},\ldots,1_{k_{l}},0_{k_{l}+1},\ldots,0_{n_{l}}]$ where $k_{l}$ is the remaining capacity of each layer ranging in $(z_{l},n_{l}]$. Therefore, Eqn.\ref{4} can be simplified as:
\begin{equation}\label{6}
\begin{aligned}
\mathop{\arg\min}_{{\textbf{Q}'}_{l}, {\textbf{R}'}_{l}}\parallel\textbf{Q}^{T}_{l}\textbf{y}_{l}-{\textbf{Q}'}^{T}_{l}(\textbf{m}_{l}\odot\gamma({\textbf{R}'}_{l}^{T}({\textbf{Q}'}_{l-1}^{T}{\textbf{y}'}_{l-1})))\parallel_{F}^{2}
\end{aligned}
\end{equation}
Eqn.\ref{6} is more efficient to solve by gradient descent optimizers once $\textbf{m}_{l}$ is specified in advance.

\subsection{Implementation Details}

Different from other methods, as shown in Fig.\ref{multi_layers}, our method requires layer decomposition before pruning to obtain each layer's embedding space. In the following experiments, we first apply SVD cross-channel decomposition to weight matrices, as is introduced in supplementary materials. We decompose $k\times k$ kernels into $k\times k$ and $1\times 1$, which serve as an embedding matrix and transformation matrix respectively. 

To further estimate the valid capacity of each layer, we lower the rank of each layer as much as possible during SVD decomposition under the guarantee of barely accuracy drop (its another attached benefit is the reduction of optimization parameter space of the pruned layers). According to the information propagation characteristic of ReLU, the valid number of output neurons is at least greater than this rank. Our implementation procedures are summarized as Algorithm 1 in our supplementary materials.

\section{Evaluation and Results}\label{experiments}

\subsection{Benchmark Datasets}
In this section, we mainly conduct our experiments on CIFAR-10 \cite{cifar} and ILSVRC-2012 \cite{Russakovsky2015ImageNet} to demonstrate the less information loss property of our proposed method. 
\begin{itemize}
\item{\bfseries CIFAR-10} is a dataset for 10-categories image classification, consisting of 50k color images for training and 10k color images for testing with resolution of $32\times 32$. For this experiment, we adopt VGG-9 \cite{VGG9} to compare our results with another state-of-the-art neuron pruning method, and analyze the advantages of our method in detail. 
\item{\bfseries ImageNet} is a large scale image classification dataset, which comprises 1.28 million images from 1000 categories for training and 50k images for validation. For this experiment, we adopt two of the most representative networks, i.e., VGG-16 \cite{Simonyan2014Very} and ResNet-50 \cite{He2016Deep} to demonstrate our method's generalization to large scale networks.
\end{itemize}

\subsection{VGG-9 on CIFAR-10}
VGG-9 \cite{VGG9} is derived from VGGNet \cite{Simonyan2014Very} and its architecture can be denoted as "$(2\times 64C3)-MP2-(2\times 128C3)-MP2-(2\times 256C3)-MP2-(2\times 512FC)-10/100FC-Softmax$". In order to improve its convergence, a batch normalization layer is appended after each convolutional layer. We train this network with SGD optimization method with mini-batch size 100, weight decay 1e-4 and momentum 0.9. Training is started by a learning rate 0.1 with linear decaying policy, and is stopped after 100k iterations. For data augmentation, only horizontal flipping and random cropping are adopted. 92\% accuracy is achieved as our baseline after training. After layer-by-layer pruning and reconstruction, the slimmer network is fine-tuned with 10k iterations by a learning rate 0.01 with linear decaying policy. In the following experiments, speed-up means theoretical speed-up, namely FLOPs reduction in this paper.

\begin{table*}[ht]
\newcommand{\tabincell}[2]{\begin{tabular}{@{}#1@{}}#2\end{tabular}}
\centering
\caption{Rank analysis of VGG-9 on CIFAR-10 and four different speed-up settings. (Num. denotes the number of output feature maps, and Spar. denotes the sparsity of weight matrices. Both the filter level and channel level sparsity are considered.)}
\label{speed-up settings}
\begin{tabular}{c|c|c|c|c|c|c|c|c|c|c}
\hline
\multirow{2}{*}{Layers} & \multirow{2}{*}{\tabincell{c}{Num.}} & \multirow{2}{*}{\tabincell{c}{Low\\rank}} &\multicolumn{2}{|c|}{$2\times$speed-up} & \multicolumn{2}{|c|}{$3\times$speed-up}& \multicolumn{2}{|c|}{$4\times$speed-up}& \multicolumn{2}{|c}{$5\times$speed-up}\\ 
\cline{4-11}
&&&Num.&Spar.&Num.&Spar.&Num.&Spar.&Num.&Spar.\\
\hline
CONV1\_1&64&6&12&81.3\%&6&90.6\%&6&90.6\%&6&90.6\%\\
CONV1\_2&64&18&36&89.5\%&18&97.4\%&18&97.4\%&18&97.4\%\\
CONV2\_1&128&37&74&67.5\%&65&85.7\%&37&91.9\%&37&91.9\%\\
CONV2\_2&128&49&98&55.7\%&98&61.1\%&69&84.4\%&49&88.9\%\\
CONV3\_1&256&89&236&29.4\%&178&46.8\%&178&62.5\%&152&77.3\%\\
CONV3\_2&256&103&256&7.8\%&206&44.0\%&206&44.0\%&206&52.2\%\\
\hline
\end{tabular}
\end{table*}

Since almost all computational cost lies in convolutional layers instead of fully-connected layers, we only prune the preceding six convolutional layers of VGG-9 in this experiment in order to reach largest speed-up. Before that, we apply SVD decomposition to these layers and analyze the low ranks of them. We find that when we lower the preserving cumulative sum of largest singular values in each layer to 0.55, the accuracy of the decomposed VGG-9 can be exactly recovered through short-term fine-tuning and the ranks are shown in Tab. \ref{speed-up settings}. Based on the discussion about the valid number of neurons in the above section, we can assign a proper pruning ratio range to each layer. As shown in Tab.\ref{speed-up settings}, we design four pruning configurations for four different speed-up demands. Considering the reconstruction error in shallow layers can be compensated by deeper layers and not vice versa, we assign lower sparsity to deeper layers. 

\begin{figure}[tp]
\centering
\includegraphics[scale=0.2]{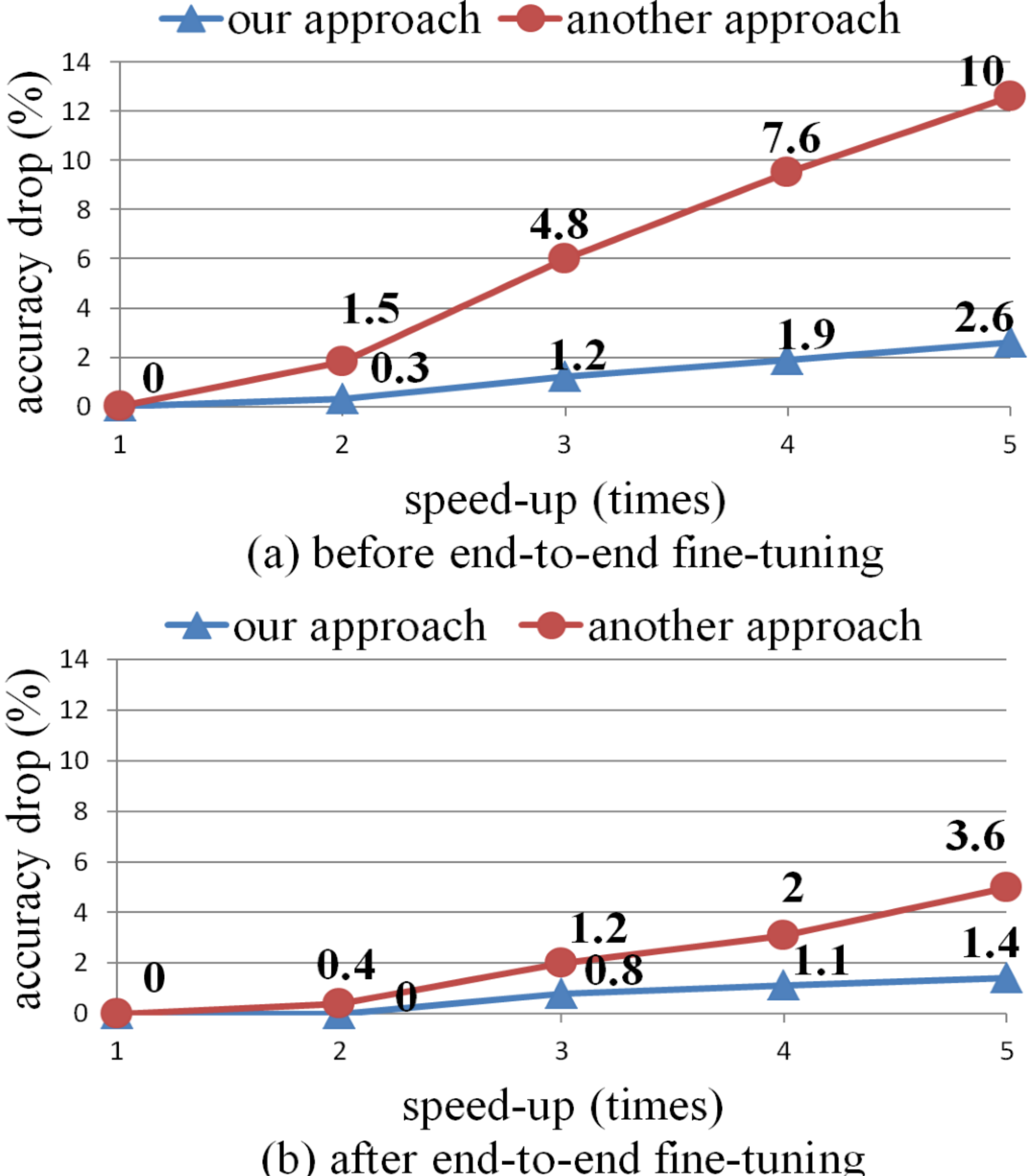}
\caption{Performance comparison of pruned VGG-9 on CIFAR-10 between our method and another state-of-the-art neuron pruning method \cite{He2017Channel} with four different speed-up configurations.}
\label{compason_cifar10}
\end{figure}

According to these configurations, we apply neuron pruning to VGG-9 by our proposed LDRF. Before that, we do comparison experiments with several different neuron selection methods \cite{DBLP:journals/corr/HuPTT16,Molchanov2016Pruning} to demonstrate our framework's insensitivity to neuron selection methods as discussed in above session. In Tab.\ref{Neuron selection methods}, each result is comparable. So we simply use Top-k to initialize $\textbf{m}$ to solve Eqn.\ref{6} in the following experiments. The accuracy is recovered under $2\times$ speed-up, and only suffers from 1.4\% drop under $5\times$ speed-up. To further verify the efficiency of our method, we implement the channel pruning method of He \emph{et al.} \cite{He2017Channel} with the same pruning configurations, which adopts a layer-by-layer optimization mode similar to Fig.\ref{layerwise}. From Fig.\ref{compason_cifar10}, before end-to-end fine-tuning, our method suffers from significantly less accuracy drop owing to the less information loss during propagation. Our superiority is more apparent as the acceleration rate increases. We can achieve better performance after fine-tuning since we provides a better initiation point for fine-tuning.

\begin{table}[tp]
\newcommand{\tabincell}[2]{\begin{tabular}{@{}#1@{}}#2\end{tabular}}
\centering
\caption{Performance comparison of pruned VGG-9 (5$\times$) on CIFAR-10 with different heuristic neuron selection methods before end-to-end fine-tuning.}
\label{Neuron selection methods}
\begin{tabular}{c|c|c|c|c|c}
\hline
Method&Top-k&Random&APoZ&Activation&Weight\\
\hline
Error$\uparrow$&2.61\%&2.81\%&2.72\%&2.57\%&2.76\%\\
\hline
\end{tabular}
\end{table}
\begin{figure}[tp]
\centering
\includegraphics[scale=0.24]{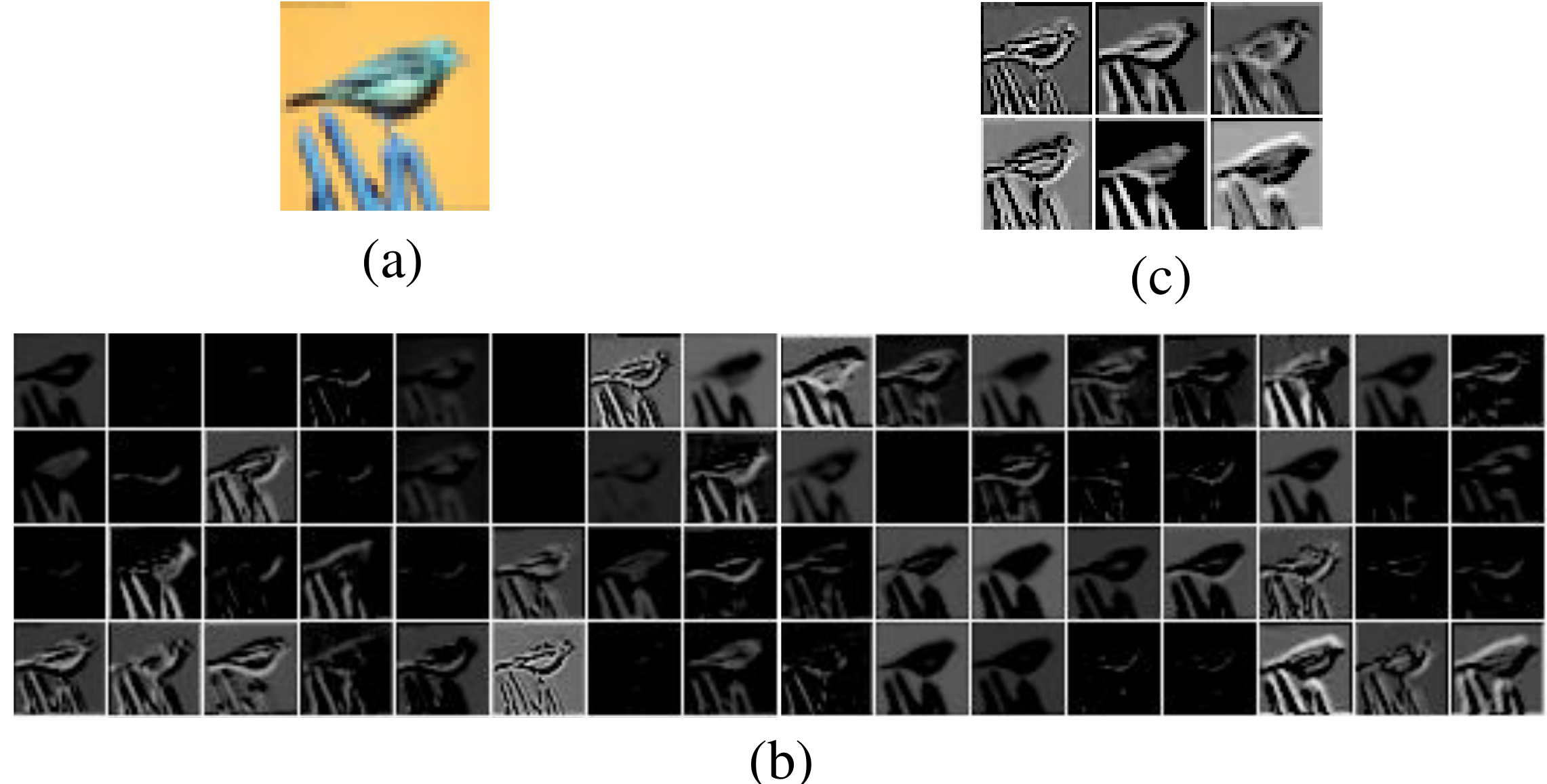}
\caption{Feature maps visualization. (a) is an image from CIFAR-10. (b) are the redundant output feature maps from the first layer of VGG-9 before pruning. (c) are the informative output feature maps from the first layer reconstructed by our proposed method. (All feature maps are non-linearly enhanced for more clear visualization)}
\label{vis_cifar10}
\end{figure}

To further comfirm the impact of our method, we select an image from CIFAR-10, and visualize the feature maps of CONV1\_1 before pruning and our reconstructed feature maps after pruning in Fig.\ref{vis_cifar10}. Despite the redundancy in the original feature maps, it is impossible to select a fraction of feature maps from them, which involve the entire information, to reconstruct the output of the next layer. Our method provides a direction to squeeze the entire information into a fraction of feature maps as much as possible and propagate to the following layers. As shown in Fig.\ref{vis_cifar10}, our reconstructed feature maps are more informative than any fractions selected from the original feature maps.

\subsection{VGG-16 and ResNet-50 on ImageNet-2012}

To demonstrate the generalization ability of our method to large scale networks, we adopt two most representative networks, i.e., VGG-16 and ResNet-50 to compare our method with other recently proposed state-of-the-art pruning methods. We conduct our experiments on ImageNet-2012 benchmark. We scale the short-side of images to 256 and adopt $224\times 224$ random crop as well as horizontal flip to augment the training dataset. At the validation phase, we only center crop the feeding resized images to $224\times 224$. We present our results by top-5 validation accuracy of single-view approach. The single-view top-5 accuracy of the baseline network is 89.9\% and 92.2\% for VGG-16 and ResNet-50 respectively. To further evaluate the performance, we use Caffe framework to test the practical speed-ups of each pruned network on GPU (TITAN X Pascal, CUDA8 and cuDNN5) and CPU (Intel Xeon E5-2650, atlas).

{\bfseries VGG-16} is a redundant network with nearly 90\% parameters stored in fully-connected layers. To reduce its storage overhead is an easier task already solved by other methods, like Han \emph{et al.} \cite{Han2015Learning,DeepCompression}. However, almost 97\% computational cost comes from its convolutional layers. Pruning convolutional layers is more difficult than pruning  fully-connected layers. Hence, we mainly focus on pruning convolutional layers just like other state-of-the-art neuron pruning methods \cite{Li2016Pruning,He2017Channel,Luo2017ThiNet}.
\begin{figure}[tp]
\centering
\includegraphics[scale=0.33]{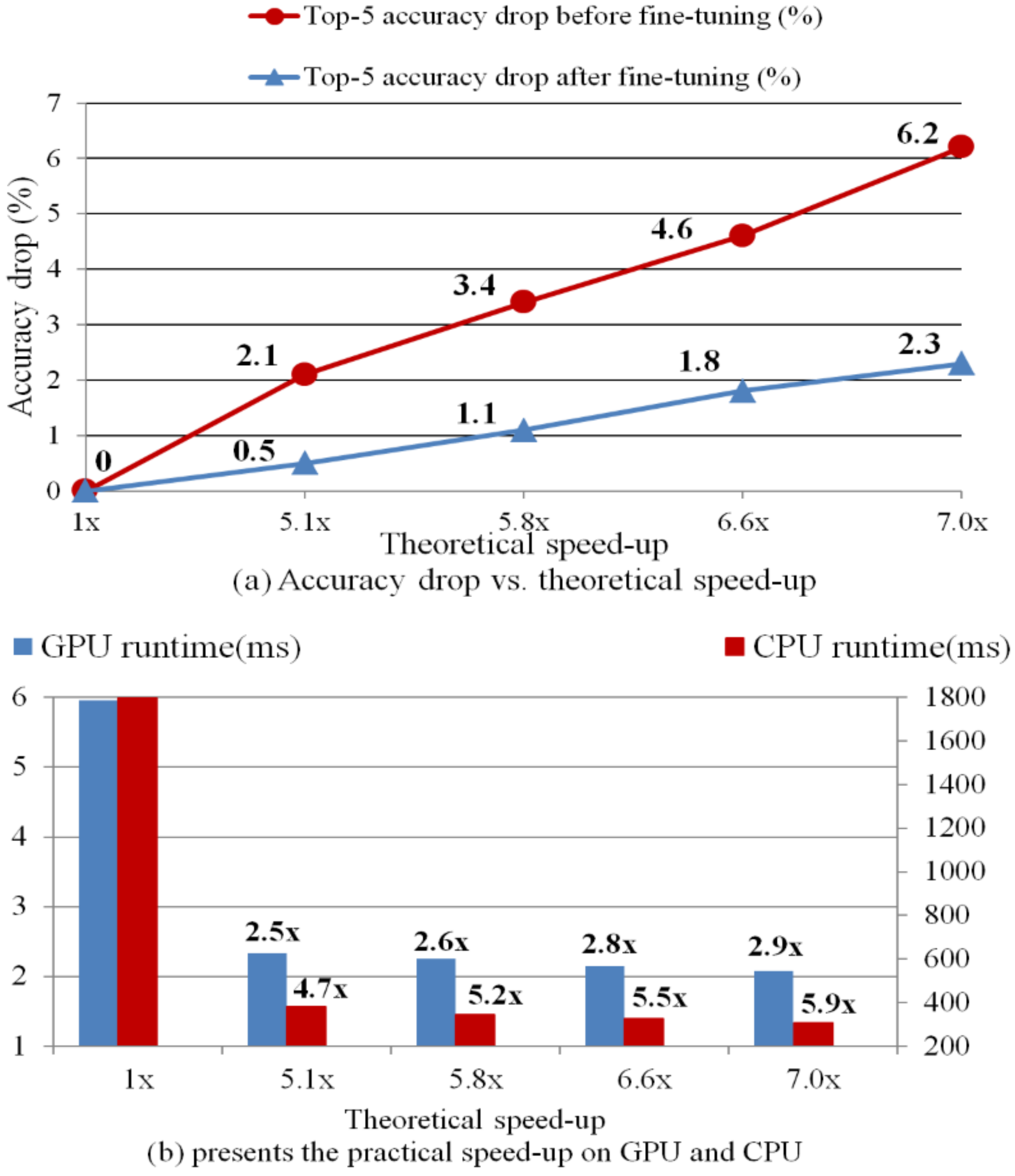}
\caption{The results of VGG-16 on ImageNet with four different speed-up configurations, as well as the comparison between theoretical and practical speed-up.}
\label{vgg16_results}
\end{figure}
\begin{table}[tp]
\newcommand{\tabincell}[2]{\begin{tabular}{@{}#1@{}}#2\end{tabular}}
\centering
\caption{Performance of pruned VGG-16 on ImageNet compared with other neuron pruning methods.}
\label{Channel Pruning VGG-16}
\begin{tabular}{c|c|c|c}
\hline
Method& \tabincell{c}{Top-5\\(1-view)} &\tabincell{c}{Increased\\error} &Speed-up\\
\hline
Original&89.9\%&0.0\%&1.00$\times$\\
Molchanov \emph{et al.}-1&87.0\%&2.9\%&2.69$\times$\\
Molchanov \emph{et al.}-2&84.5\%&5.4\%&3.87$\times$\\
Luo \emph{et al.}&89.5\%&0.4\%&3.22$\times$\\
He \emph{et al.}-1&88.9\%&1.0\%&4.00$\times$\\
He \emph{et al.}-2&88.2\%&1.7\%&5.00$\times$\\
\hline
LDRF (Ours)&{\bfseries89.4\%}&{\bfseries0.5\%}&{\bfseries5.13$\times$}\\
\hline
\end{tabular}
\end{table}
\begin{table}[tp]
\newcommand{\tabincell}[2]{\begin{tabular}{@{}#1@{}}#2\end{tabular}}
\centering
\caption{Performance of pruned ResNet-50 on ImageNet compared with other neuron pruning methods.}
\label{ResNet-50}
\begin{tabular}{c|c|c|c}
\hline
Method& \tabincell{c}{Top-5\\(1-view)}&\tabincell{c}{Increased\\error}&Speed-up\\
\hline
Original&92.20\%&0.00\%&1.00$\times$\\
He \emph{et al.}&90.80\%&1.40\%&2.00$\times$\\
Luo \emph{et al.}-1&90.67\%&1.53\%&1.58$\times$\\
Luo \emph{et al.}-2&90.02\%&2.18\%&2.26$\times$\\
\hline
LDRF (Ours)-1&{\bfseries92.01\%}&{\bfseries0.19\%}&{\bfseries2.27$\times$}\\
LDRF (Ours)-2&{\bfseries91.55\%}&{\bfseries0.65\%}&{\bfseries3.00$\times$}\\
\hline
\end{tabular}
\end{table}
\begin{table}[tp]
\newcommand{\tabincell}[2]{\begin{tabular}{@{}#1@{}}#2\end{tabular}}
\centering
\caption{The comparison between an existing method and our method before and after end-to-end fine-tuning (FT).}
\label{compare}
\begin{tabular}{c|c|c|c}
\hline
\tabincell{c}{Network\\($5\times$speed-up)}& FT &\tabincell{c}{Error $\uparrow$ \\ (He \emph{et al.})} & \tabincell{c} {Error $\uparrow$ \\ (Ours)}\\ 
\hline
\multirow{2}{*}{\tabincell{c}{VGG9 \\ (Cifar-10)}} & $\xmark$ &10.0\% (Our impl.)&2.6\%\\
\cline{2-4}
&$\cmark$&3.6\% (Our impl.)&1.4\%\\
\hline
\multirow{2}{*}{\tabincell{c}{VGG16 \\ (ImageNet)}} & $\xmark$ &22.0\%&2.1\%\\
\cline{2-4}
&$\cmark$&1.7\%&0.5\%\\
\hline
\end{tabular}
\end{table}

We apply cross-channel decomposition to each convolutional layers through SVD to obtain each layer's embedding space. During this process, the preserving cumulative sum of largest singular values can be lowered to 0.65 while keeping the accuracy unchanged. The low rank of each layer is presented in supplementary materials which provides us a proper pruning ratio range. Accordingly, we assign 4 different pruning ratios in this range to meet four different speed-up demands ($5.1\times$, $5.8\times$, $6.6\times$, $7.0\times$). For each layer's reconstruction, it begins converging with about 0.1 epoch training. The experiment results after pruning and optimization are shown in Fig.\ref{vgg16_results}(a), where the error increases accordingly as the pruning ratio increases. Our method can speed the network up to $5.1\times$ with only 0.5\% top-5 accuracy drop, which is significantly better than other neuron pruning methods as shown in Tab.\ref{Channel Pruning VGG-16}. In Fig.\ref{vgg16_results}(b), there is a gap between practical and theoretical speed-up. The incremental computation efficiency reflected on CPU is more than that on GPU, since GPU is a memory-bound computational platform. It means FLOPs reduction is not a complete metric to measure practical runtime. I/O delay or other factors should be taken into consideration which will be studied in our future works.

{\bfseries ResNet-50} is a very deep network which is more challenging than VGG-16 for neuron pruning. The information loss accumulation during layer-by-layer pruning and reconstruction will keep growing as layer increases. Notably, our proposed method benefits to avoid such problem owing to the lossless information propagation characteristic. In line with the procedures to VGG-16, our method can speed ResNet-50 up to about $2.27\times$ and $3\times$ with only 0.19\% and 0.65\% top-5 accuracy drop respectively. In Tab.\ref{ResNet-50}, our results outperform other state of the arts by a large margin. 

\subsection{Training Burden Discussion}
Cross-channel decomposition is the only extra burden compared with the existing layer-by-layer pruning and reconstruction methods, which is only executed once before pruning and can be viewed as a preprocessing to generate embedding space and analyse the channel redundancy for each layer (note that pruning ratio assigned to each layer denotes the remaining capacity and determines the ceiling of network accuracy). It costs us several epochs training to do such preprocessing. But it is a worthy cost to get a much better results. For a fair comparison, we compare our results before end-to-end fine-tuning with the results of an existing method after fine-tuning. As shown in Tab.\ref{compare}, our results before end-to-end fine-tuning are comparable with the results of the existing method after fine-tuning. Our method has a much greater potential to achieve a much better result after fine-tuning. The essence of our good results is from the less information loss property of our proposed framework.

\section{Conclusion}

In this paper we propose a novel neuron pruning method that squeezes intact information into survived neurons instead of discarding unimportant neurons abruptly to prevent from information loss. Signals before and after pruning are projected into an shared embedding space which serves as a function of information buffer. Because of less information loss accumulation compared with other neuron pruning methods, our method achieves superior results, which is more apparent before end-to-end fine-tuning. 



\bibliographystyle{aaai}
\bibliography{aaai}

\begin{thebibliography}{}

\bibitem[\protect\citeauthoryear{Aghasi, Nguyen, and Romberg}{2017}]{Net-Trim}
Aghasi, A.; Nguyen, N.; and Romberg, J.~K.
\newblock 2017.
\newblock Net-trim: Convex pruning of deep neural networks with performance
  guarantee.
\newblock In {\em Advances in Neural Information Processing Systems}.

\bibitem[\protect\citeauthoryear{Alvarez and
  Salzmann}{2016}]{alvarez2016learning}
Alvarez, J.~M., and Salzmann, M.
\newblock 2016.
\newblock Learning the number of neurons in deep networks.
\newblock In {\em Advances in Neural Information Processing Systems}.

\bibitem[\protect\citeauthoryear{Anwar, Hwang, and
  Sung}{2017}]{anwar2017structured}
Anwar, S.; Hwang, K.; and Sung, W.
\newblock 2017.
\newblock Structured pruning of deep convolutional neural networks.
\newblock {\em ACM Journal on Emerging Technologies in Computing Systems}
  13(3):32.

\bibitem[\protect\citeauthoryear{Courbariaux, Bengio, and David}{2015}]{VGG9}
Courbariaux, M.; Bengio, Y.; and David, J.-P.
\newblock 2015.
\newblock Binaryconnect: Training deep neural networks with binary weights
  during propagations.
\newblock In {\em Advances in Neural Information Processing Systems}.

\bibitem[\protect\citeauthoryear{Dahl, Sainath, and
  Hinton}{2013}]{dahl2013improving}
Dahl, G.~E.; Sainath, T.~N.; and Hinton, G.~E.
\newblock 2013.
\newblock Improving deep neural networks for lvcsr using rectified linear units
  and dropout.
\newblock In {\em International Conference on Acoustics, Speech, and Signal
  Processing}.

\bibitem[\protect\citeauthoryear{Guo, Yao, and Chen}{2016}]{Guo2016Dynamic}
Guo, Y.; Yao, A.; and Chen, Y.
\newblock 2016.
\newblock Dynamic network surgery for efficient dnns.
\newblock In {\em Advances in Neural Information Processing Systems}.

\bibitem[\protect\citeauthoryear{Han \bgroup et al\mbox.\egroup
  }{2015}]{Han2015Learning}
Han, S.; Pool, J.; Tran, J.; and Dally, W.~J.
\newblock 2015.
\newblock Learning both weights and connections for efficient neural networks.
\newblock In {\em Advances in Neural Information Processing Systems}.

\bibitem[\protect\citeauthoryear{Han, Mao, and Dally}{2016}]{DeepCompression}
Han, S.; Mao, H.; and Dally, W.~J.
\newblock 2016.
\newblock Deep compression: Compressing deep neural network with pruning,
  trained quantization and huffman coding.
\newblock In {\em International Conference on Learning Representations}.

\bibitem[\protect\citeauthoryear{Hassibi and Stork}{1993}]{Hassibi1993Second}
Hassibi, B., and Stork, D.~G.
\newblock 1993.
\newblock Second order derivatives for network pruning: Optimal brain surgeon.
\newblock In {\em Advances in Neural Information Processing Systems}.

\bibitem[\protect\citeauthoryear{He \bgroup et al\mbox.\egroup
  }{2016}]{He2016Deep}
He, K.; Zhang, X.; Ren, S.; and Sun, J.
\newblock 2016.
\newblock Deep residual learning for image recognition.
\newblock In {\em Proceedings of the IEEE Conference on Computer Vision and
  Pattern Recognition}.

\bibitem[\protect\citeauthoryear{He, Zhang, and Sun}{2017}]{He2017Channel}
He, Y.; Zhang, X.; and Sun, J.
\newblock 2017.
\newblock Channel pruning for accelerating very deep neural networks.
\newblock In {\em Proceedings of International Conference on Computer Vision}.

\bibitem[\protect\citeauthoryear{Hu \bgroup et al\mbox.\egroup
  }{2016}]{DBLP:journals/corr/HuPTT16}
Hu, H.; Peng, R.; Tai, Y.; and Tang, C.
\newblock 2016.
\newblock Network trimming: {A} data-driven neuron pruning approach towards
  efficient deep architectures.
\newblock {\em CoRR} abs/1607.03250.

\bibitem[\protect\citeauthoryear{Krizhevsky and Hinton}{2009}]{cifar}
Krizhevsky, A., and Hinton, G.
\newblock 2009.
\newblock Learning multiple layers of features from tiny images.

\bibitem[\protect\citeauthoryear{Lebedev and Lempitsky}{2016}]{Lebedev2016Fast}
Lebedev, V., and Lempitsky, V.
\newblock 2016.
\newblock Fast convnets using group-wise brain damage.
\newblock In {\em Proceedings of the IEEE Conference on Computer Vision and
  Pattern Recognition}.

\bibitem[\protect\citeauthoryear{Lecun, Denker, and
  Solla}{1990}]{lecun1990optimal}
Lecun, Y.; Denker, J.~S.; and Solla, S.~A.
\newblock 1990.
\newblock Optimal brain damage.
\newblock In {\em Advances in Neural Information Processing Systems}.

\bibitem[\protect\citeauthoryear{Li \bgroup et al\mbox.\egroup
  }{2017}]{Li2016Pruning}
Li, H.; Kadav, A.; Durdanovic, I.; Samet, H.; and Graf, H.~P.
\newblock 2017.
\newblock Pruning filters for efficient convnets.
\newblock In {\em International Conference on Learning Representations}.

\bibitem[\protect\citeauthoryear{Liu \bgroup et al\mbox.\egroup
  }{2015}]{Liu2015Sparse}
Liu, B.; Wang, M.; Foroosh, H.; and Tappen, M.
\newblock 2015.
\newblock Sparse convolutional neural networks.
\newblock In {\em Proceedings of the IEEE Conference on Computer Vision and
  Pattern Recognition}.

\bibitem[\protect\citeauthoryear{Liu \bgroup et al\mbox.\egroup
  }{2017}]{slimming}
Liu, Z.; Li, J.; Shen, Z.; Huang, G.; Yan, S.; and Zhang, C.
\newblock 2017.
\newblock Learning efficient convolutional networks through network slimming.
\newblock In {\em Proceedings of International Conference on Computer Vision}.

\bibitem[\protect\citeauthoryear{Luo, Wu, and Lin}{2017}]{Luo2017ThiNet}
Luo, J.~H.; Wu, J.; and Lin, W.
\newblock 2017.
\newblock Thinet: A filter level pruning method for deep neural network
  compression.
\newblock In {\em Proceedings of International Conference on Computer Vision}.

\bibitem[\protect\citeauthoryear{Mittal \bgroup et al\mbox.\egroup
  }{2018}]{randompruning}
Mittal, D.; Bhardwaj, S.; Khapra, M.~M.; and Ravindran, B.
\newblock 2018.
\newblock Recovering from random pruning: On the plasticity of deep
  convolutional neural networks.
\newblock In {\em Workshop on Applications of Computer Vision}.

\bibitem[\protect\citeauthoryear{Molchanov \bgroup et al\mbox.\egroup
  }{2017}]{Molchanov2016Pruning}
Molchanov, P.; Tyree, S.; Karras, T.; Aila, T.; and Kautz, J.
\newblock 2017.
\newblock Pruning convolutional neural networks for resource efficient
  inference.
\newblock In {\em International Conference on Learning Representations}.

\bibitem[\protect\citeauthoryear{Russakovsky \bgroup et al\mbox.\egroup
  }{2015}]{Russakovsky2015ImageNet}
Russakovsky, O.; Deng, J.; Su, H.; Krause, J.; Satheesh, S.; Ma, S.; Huang, Z.;
  Karpathy, A.; Khosla, A.; and Bernstein, M.
\newblock 2015.
\newblock Imagenet large scale visual recognition challenge.
\newblock {\em International Journal of Computer Vision} 115(3):211--252.

\bibitem[\protect\citeauthoryear{Sandler \bgroup et al\mbox.\egroup
  }{2018}]{mobilenetv2}
Sandler, M.; Howard, A.~G.; Zhu, M.; Zhmoginov, A.; and Chen, L.
\newblock 2018.
\newblock Inverted residuals and linear bottlenecks: Mobile networks for
  classification, detection and segmentation.
\newblock In {\em Proceedings of the IEEE Conference on Computer Vision and
  Pattern Recognition}.

\bibitem[\protect\citeauthoryear{Simonyan and
  Zisserman}{2015}]{Simonyan2014Very}
Simonyan, K., and Zisserman, A.
\newblock 2015.
\newblock Very deep convolutional networks for large-scale image recognition.
\newblock In {\em International Conference on Learning Representations}.

\bibitem[\protect\citeauthoryear{Szegedy \bgroup et al\mbox.\egroup
  }{2015}]{szegedy2015going}
Szegedy, C.; Liu, W.; Jia, Y.; Sermanet, P.; Reed, S.~E.; Anguelov, D.; Erhan,
  D.; Vanhoucke, V.; and Rabinovich, A.
\newblock 2015.
\newblock Going deeper with convolutions.
\newblock In {\em Proceedings of the IEEE Conference on Computer Vision and
  Pattern Recognition}.

\bibitem[\protect\citeauthoryear{Wen \bgroup et al\mbox.\egroup
  }{2016}]{Wen2016Learning}
Wen, W.; Wu, C.; Wang, Y.; Chen, Y.; and Li, H.
\newblock 2016.
\newblock Learning structured sparsity in deep neural networks.
\newblock In {\em Advances in Neural Information Processing Systems}.

\bibitem[\protect\citeauthoryear{Ye \bgroup et al\mbox.\egroup
  }{2018}]{ye2018rethinking}
Ye, J.; Lu, X.; Lin, Z.; and Wang, J.~Z.
\newblock 2018.
\newblock Rethinking the smaller-norm-less-informative assumption in channel
  pruning of convolution layers.
\newblock In {\em International Conference on Learning Representations}.

\bibitem[\protect\citeauthoryear{Yu \bgroup et al\mbox.\egroup
  }{2018}]{yu2018nisp:}
Yu, R.; Li, A.; Chen, C.; Lai, J.; Morariu, V.~I.; Han, X.; Gao, M.; Lin, C.;
  and Davis, L.~S.
\newblock 2018.
\newblock Nisp: Pruning networks using neuron importance score propagation.
\newblock {\em Proceedings of the IEEE Conference on Computer Vision and
  Pattern Recognition}.

\bibitem[\protect\citeauthoryear{Yuan and Lin}{2006}]{yuan2006model}
Yuan, M., and Lin, Y.
\newblock 2006.
\newblock Model selection and estimation in regression with grouped variables.
\newblock {\em Journal of The Royal Statistical Society Series B-statistical
  Methodology} 68(1):49--67.

\end{thebibliography}

\medskip
\medskip
\medskip

\section{Supplementary Materials}

{\bfseries Pipeline.} We summarize the detailed procedures of our proposed LDRF as Algorithm \ref{algorithm}. In line with Fig.3 in the paper, they are mainly divided into three stages: 1) Applying cross-channel decomposition to each layer to obtain the corresponding embedding space. Simultaneously, we estimate the low-rank of each layer under the guarantee of barely accuracy drop, guiding us to assign a proper pruning ratio range to each layer which is crucial to the ceiling of the theoretical accuracy of pruned networks. 2) Pruning each layer while compensating the information loss of the discarded neurons in the embedding space in a layer-by-layer manner. 3) Recomposing the adjacent linear weight matrices to avoid depth increasing, since depth increase will lead to IO delay.

\begin{algorithm}[tp]
\caption{The implementation details of Layer Decomposition-Recomposition Framework (LDRF) for neuron pruning.}\label{algorithm}
{\bf Input:}
A given CNN network $\mathcal{N}$ with $\{\textbf{W}_{1}$,...,$\textbf{W}_{L}\}$ trained by a training dataset with labels $\{\mathcal{X},\mathcal{T}\}$
\begin{algorithmic}[1]
\State //Stage 1: solve Eqn.[6]
\State //$e$ is the preserving cumulative sum of largest singular values for rank reduction
\State //$\varepsilon$ can be fixed as a small value or be set by the users.
\For{$e=[\varepsilon:0.05:0.9]$}
\State Apply SVD to $\{\textbf{W}_{1}$,...,$\textbf{W}_{L}\}$ to decompose $k\times k$ kernels into $k\times k$ (\textbf{Q}) and $1\times 1$ (\textbf{R})
\State Short-term fine-tuning with $\{\mathcal{X},\mathcal{T}\}$
\If{recover the accuracy of $\mathcal{N}$}
\State Obtain $\{\textbf{Q}_{0}$,...,$\textbf{Q}_{L-1}$, $\textbf{R}_{1}$,...,$\textbf{R}_{L}\}$ and $\{z_{1}$,...,$z_{L}\}$
\State break
\EndIf
\EndFor
\State //Stage 2: solve Eqn.[7]
\State $\textbf{y}_{0}\xleftarrow{}\mathcal{X}$, $\textbf{y}'_{0}\xleftarrow{}\mathcal{X}$
\State $\{\textbf{Q}'_{0}$,...,$\textbf{Q}'_{L-1}$, $\textbf{R}'_{1}$,...,$\textbf{R}'_{L}\}\xleftarrow{}\{\textbf{Q}_{0}$,...,$\textbf{Q}_{L-1}$, $\textbf{R}_{1}$,...,$\textbf{R}_{L}\}$
\For{$l=[1:1:L-1]$}
\State $\textbf{m}_{l}\xleftarrow{}[1_{1},\ldots,1_{k_{l}},0_{k_{l}+1},\ldots,0_{n_{l}}],k_{l}\in(z_{l},n_{l}]$
\State $\textbf{y}_{l}\xleftarrow{}\gamma(\textbf{R}_{l}^{T}\textbf{Q}_{l-1}^{T}\textbf{y}_{l-1})$
\State ${\textbf{Q}'}_{l}, {\textbf{R}'}_{l}\xleftarrow{}\mathop{\arg\min}_{{\textbf{Q}'}_{l}, {\textbf{R}'}_{l}}\parallel\textbf{Q}^{T}_{l}\textbf{y}_{l}-{\textbf{Q}'}^{T}_{l}(\textbf{m}_{l}\odot\gamma({\textbf{R}'}_{l}^{T}({\textbf{Q}'}_{l-1}^{T}{\textbf{y}'}_{l-1})))\parallel_{F}^{2}$
\State $\textbf{y}'_{l}\xleftarrow{}\gamma({\textbf{R}'}_{l}^{T}{\textbf{Q}'}_{l-1}^{T}\textbf{y}'_{l-1})$
\EndFor
\State ${\textbf{R}'}_{L}\xleftarrow{}\mathcal{L}({\textbf{R}'}_{L}^{T}{\textbf{Q}'}_{L-1}^{T}{\textbf{y}'}_{L-1}\,\,\vert\,\,\mathcal{T})$, $\mathcal{L}$ is a loss function
\State //Stage 3: Recompose $\textbf{Q}'$ and $\textbf{R}'$ to obtain a compact transformation matrix $\textbf{W}'$
\For{$l=[1:1:L]$}
\State $\textbf{W}'_{l}=\textbf{Q}'_{l-1}\textbf{R}'_{l}$
\State Removing useless filters according to $\textbf{m}_{l}$
\EndFor
\end{algorithmic}
{\bf Output:}
a slimmer network $\mathcal{N}'$ with $\{\textbf{W}'_{1}$,...,$\textbf{W}'_{L}\}$
\end{algorithm}

\begin{table}[tp]
\centering
\caption{Pruning ratio analysis for each layer in VGG-16. The range in this table denotes the range of the valid neuron number of each layer. The left endpoint of each range is the low-rank estimation after rank reduction through SVD, while the right endpoint is the original neuron number of each layer. When pruning ratio is assigned to each layer, we should at least keep the number of preserving neurons locating in this range, or it is bound to lose the flow-in information according to the discussion of the information propagation characteristic of ReLU layer in the main body.}
\label{vgg16}
\begin{tabular}{c|ccc}
\hline
Layer & CONV1\_1 & CONV1\_2 & CONV2\_1\\ 
\hline
Range&[4,64]&[14,64]&[28,128]\\
\hline
Layer  & CONV2\_2 & CONV3\_1 & CONV3\_2 \\
\hline
Range &[44,128]&[72,256]&[85,256]\\
\hline
Layer & CONV3\_3 & CONV4\_1 & CONV4\_2 \\
\hline
Range &[101,256]&[167,512]&[207,512]\\
\hline
Layer& CONV4\_3 & CONV5\_1 & CONV5\_2\\ 
\hline
Range&[231,512]&[236,512]&[234,512]\\
\hline
Layer & CONV5\_3 & &\\
\hline
Range &[234,512] & & \\
\hline
\end{tabular}
\end{table}

{\bfseries SVD cross-channel decomposition.} We decompose $k\times k$ filters into $k\times k$ and $1\times 1$ ones through SVD as follows:
\begin{equation}
\textbf{W}_l=\textbf{Q}_l\textbf{R}_l
\end{equation}
where the shape of $\textbf{W}_l$ is $(k, k, c, n)$, $c$ and $n$ denote the input and output channels, respectively. We firstly reshape it into 2D matrix $(k\times k\times c, n)$, and then apply SVD to decompose it into $(k\times k\times c, n\_rank)$ and $(n\_rank, n)$ matrices, and finally remap it back to two tensors with shape $(k, k, c, n\_rank)$ and $(1, 1, n\_rank, n)$. 

$\textbf{Q}_{l}^{T}\textbf{y}_{l-1}$ is the embedding representation used for layer reconstruction. In order to simplify the optimization, it is a better choice to normalize the scale of $\textbf{Q}_{l}^{T}\textbf{y}_{l-1}$ so that only one learning rate is enough to control the learning of each layer's reconstruction simultaneously. To make this feasible, we first compute the mean $m$ and variance $var$ of the embedding representation from several batches of $\textbf{Q}_{l}^{T}\textbf{y}_{l-1}$. Then the normalized embedding representation becomes:
\begin{equation}
\widehat{\textbf{Q}_{l}^{T}\textbf{y}_{l-1}} =  \frac{\textbf{Q}_{l}^{T}\textbf{y}_{l-1} - m}{\sqrt{var}}
\end{equation}
which is more efficient for layer reconstruction. Therefore, the process of decomposition is formulated as:
\begin{equation}
\!\!\textbf{W}_{l}^{T}\textbf{y}_{l\!-\!1}\!=\!\textbf{R}_{l}^{T}\textbf{Q}_{l}^{T}\textbf{y}_{l\!-\!1}\!=\!\sqrt{var}\textbf{R}_{l}^{T}(\frac{\textbf{Q}_{l}^{T}\textbf{y}_{l\!-\!1}\!-\!m}{\sqrt{var}})\!+\!m\textbf{R}^{T}
\end{equation}
In this way, the original embedding matrix $\textbf{Q}_l$ is converted to $\frac{\textbf{Q}_{l}}{\sqrt{var}}$ with bias $\frac{-m}{\sqrt{var}}$, while the transformation matrix $\textbf{R}_l$ is converted to $\sqrt{var}\textbf{R}$ with bias $m\textbf{R}$.

{\bfseries Pruning strategy.} The low-rank of each transformation matrix numerates the cross-channel redundancy. We present the pruning ratio range for each layer in VGG-16 network trained by ImageNet-2012 as shown in Table \ref{vgg16}. Generally, we can design different pruning configurations within these ranges according to users' speed-up demands. From the ranks presented in Table \ref{vgg16}, it is obvious that the shallow layers are more redundant than deeper layers. Moreover, the reconstruction error produced by shallow layers can be compensated by deeper layers and not vice versa. All these considerations guide us to prune more aggressive for shallow layers.

\end{document}